\documentclass[conference]{IEEEtran}
\IEEEoverridecommandlockouts

\usepackage{cite}
\usepackage{amsmath,amssymb,amsfonts}
\usepackage{algorithmic}
\usepackage{graphicx}
\graphicspath{{imgs/}}
\usepackage{tabularx}
\usepackage{textcomp}
\usepackage{xcolor}
\def\BibTeX{{\rm B\kern-.05em{\sc i\kern-.025em b}\kern-.08em
    T\kern-.1667em\lower.7ex\hbox{E}\kern-.125emX}}
\begin{document}

\title{DRDM: A Disentangled Representations Diffusion Model for Synthesizing Realistic Person Images\\
\footnotesize \thanks{This research is supported by the National Natural Science Foundation of China (62262045). Corresponding Email: faliang.huang@gmail.com}
}

\author{
    \IEEEauthorblockN{Enbo Huang\IEEEauthorrefmark{1}, Yuan Zhang\IEEEauthorrefmark{2}, Faliang Huang\IEEEauthorrefmark{1}, Guangyu Zhang\IEEEauthorrefmark{3}, Yang Liu\IEEEauthorrefmark{4}}
    \IEEEauthorblockA{
        \IEEEauthorrefmark{1}Guangxi Key Lab of Human-machine Interaction and Intelligent Decision, Nanning Normal University, Nanning, China\\
    }
    \IEEEauthorblockA{
        \IEEEauthorrefmark{2}Zhidayuan AI Lab, Nanning, China\\
    }
    \IEEEauthorblockA{
        \IEEEauthorrefmark{3}College of Mathematics and Informatics, South China Agricultural University, Guangzhou, China\\
    }
    \IEEEauthorblockA{
        \IEEEauthorrefmark{4}School of Computer Science and Engineering, Sun Yat-sen University, Guangzhou, China\\
    }
}

\maketitle

\begin{abstract}
Person image synthesis with controllable body poses and appearances is an essential task owing to the practical needs in the context of virtual try-on, image editing and video production.
However, existing methods face significant challenges with details missing, limbs distortion and the garment style deviation.
To address these issues, we propose a Disentangled Representations Diffusion Model (DRDM) to generate photo-realistic images from source portraits in specific desired poses and appearances.
First, a pose encoder is responsible for encoding pose features into a high-dimensional space to guide the generation of person images.
Second, a body-part subspace decoupling block (BSDB) disentangles features from the different body parts of a source figure and feeds them to the various layers of the noise prediction block, thereby supplying the network with rich disentangled features for generating a realistic target image.
Moreover, during inference, we develop a parsing map-based disentangled classifier-free guided sampling method, which amplifies the conditional signals of texture and pose.
Extensive experimental results on the Deepfashion dataset demonstrate the effectiveness of our approach in achieving pose transfer and appearance control.
\end{abstract}


\section{Introduction}
\label{sec:introduction}

\begin{figure}[ht]
	\centering
	\includegraphics[width=89mm,height=54mm]{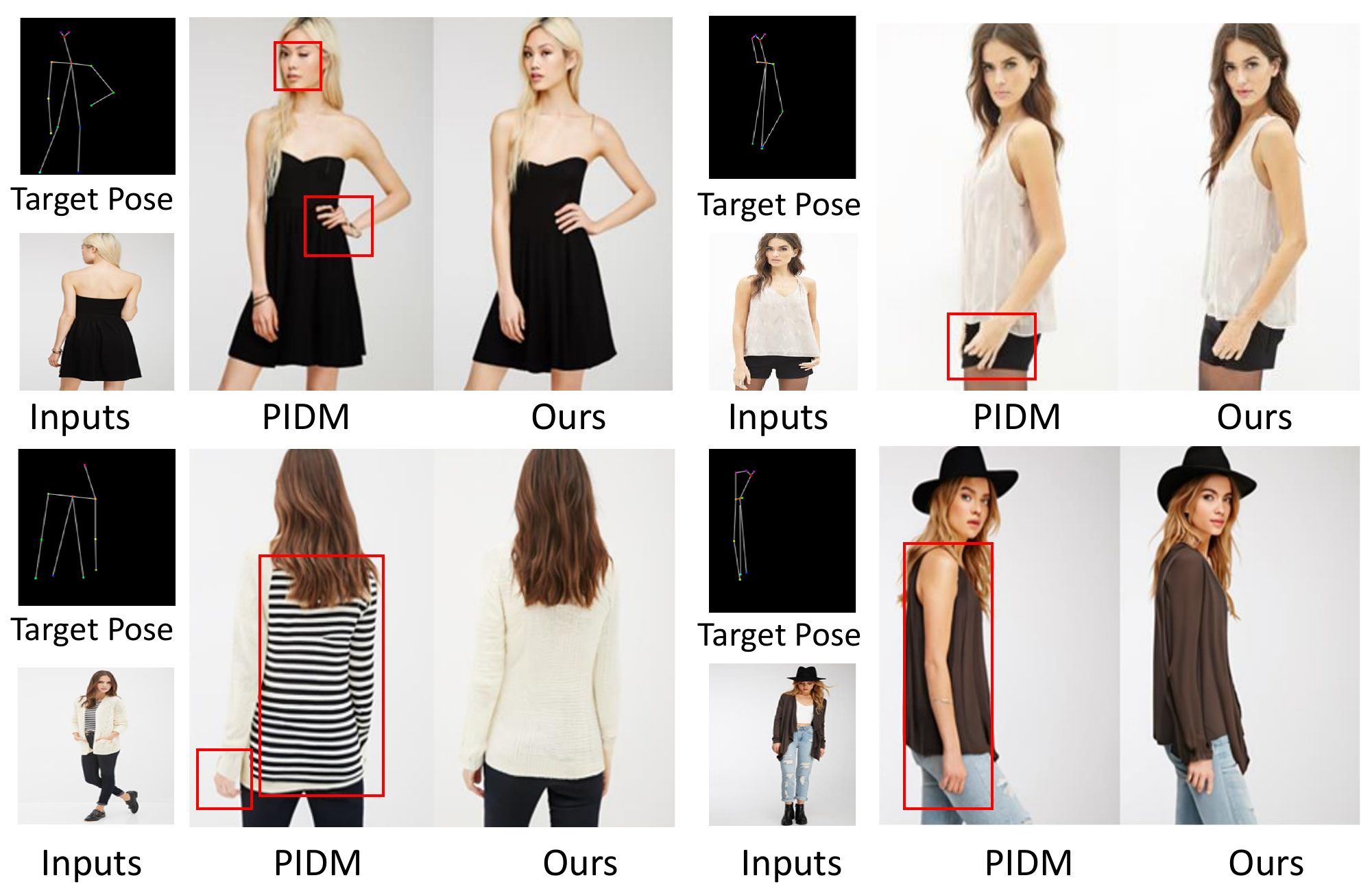}
	\vspace{-6.5mm}
	\caption{\textbf{Illustration of challenging instances and our results.} Details missing, limbs distortion and garment style deviation are three types of degeneration key issues in person image composition task. Our model can generate more accurate human limbs, clothing styles and more vivid details compared with recent state-of-the-art methods (~\emph{e.g.}, PIDM~\cite{bhunia2023person}) under variance viewpoint changed and occluded conditions.
	}
	\vspace{-2.5mm}
	\label{fig:01}
\end{figure}

The goal of the controllable person image synthesis task is to explicitly manipulate the body posture and appearance of the source individual picture, generating a photo-realistic person image that aligns with user expectations~\cite{bhunia2023person,lu2024coarse, Ju2023humansd}.
At present, majority of the approaches utilize conditional Generative Adversarial Networks (GANs) to transform the source individual photo into the desired target person image.
For example, MUST~\cite{ma2021must} and ADGAN~\cite{men2020controllable} encode style information from various semantic parts using encoders and incorporate style features through the Adaptive Instance Normalization (AdaIN) operation.
GFLA~\cite{ren2020deep} estimates the relationship between the source and target poses to guide the generation of the target photo, allowing it to preserve texture details.
By contrast, CASD~\cite{zhou2022cross} and NTED~\cite{ren2022neural} use attention mechanisms to better preserve the textural properties of the source picture.
Nevertheless, these techniques frequently yield images with unclear and indistinct details, primarily because GANs struggle to accurately establish a direct transformation pathway from the source to the target image~\cite{karras2020analyzing,karras2019style,wang2018high,karras2018progressive,zhu2017unpaired,isola2017image,ding2020ccgan}.
Recently, the diffusion model~\cite{ho2020denoising} has served as a source of inspiration for researchers to produce high-quality synthetic images by using a denoising score matching mechanism.
For instance, models such as PIDM~\cite{bhunia2023person}, OOTD~\cite{choi2024improving}, CFLD~\cite{lu2024coarse}, and HumanSD~\cite{Ju2023humansd}, etc. can effectively preserve the texture details of a source photo by using a diffusion-based approach.
However, these models face difficulties in fully maintaining the textural properties of every body parts because it directly encodes the texture of the entire source figure, leading to inaccuracies in clothing styles and limb textures, particularly when occlusions are present and postures are changed.
As shown in Figure~\ref{fig:01}, PIDM tends to generate stretched limbs and false style garments.
In fact, these problems are also commonly encountered in other spatial-temporal analysis and embodied AI tasks~\cite{liu2018hierarchically,liu2019deep,liu2021semantics,liu2022tcgl,9904187,zhu2022hybrid,liu2023cross,lin2023denselight,yan2023skeletonmae,wei2023visual,liu2024aligning}.


\vspace{-3.5mm}

To address the aforementioned issues, we propose a Disentangled Representations Diffusion Model (DRDM) that consists of the body-part subspace decoupling block (BSDB), pose encoder, and noise prediction block to achieve the disentanglement of the different body parts and the decoupling of structure and texture in person images.
Subsequently, we reassemble these disentangled appearance features according to the given target pose to generate a target person image.
In particular, BSDB utilizes the human paring map as a regularization factor to extract disentangle features from different body parts of the source person images, and then disentangled textural features were obtained through a series of dedicated texture encoders.
We introduce self-attention mechanisms within each encoder to address texture inaccuracies arising from imprecise semantic segmentation maps, allowing the texture encoders to focus solely on the accurate features of the corresponding body parts.
Finally, we use a cross-attention mechanism to pass these disentangled features to the different layers of the noise prediction block to supply the network with rich disentangled features for generating a realistic target person image.

DRDM effectively addresses common occlusion issues, distortion at limb extremities and deviation of the garment style through the disentangled feature representation.
This approach also enables explicit control over the source person image’s appearance and body posture by substituting texture features of different body parts of the source individual photo and pose features (Figure~\ref{fig:01}).
During the inference, we devised a disentangled classifier-free guided sampling based on parsing map, which amplifies the conditional signals of texture and pose.
This approach ensures a close alignment between the appearance of the generated person image and the target pose to preserve the realism of the generated result under varying poses.
The experimental and evaluation results demonstrate that our method achieves state-of-the-art results on the Deepfashion dataset.
Moreover, our approach generates images with more realistic details, and its superiority is confirmed through quantitative and qualitative comparisons and user studies.
Our major contributions are as follows:

\begin{itemize}
	\item A diffusion model with a disentangled regularization module for synthesizing highly realistic person images. We can reassemble the disentangled textural features according to the parsing map, and demonstrate that this regularization factor matters the performance in addressing common issues such as occlusion and distortion of limbs.
	
	\item A parsing map-based classifier-free disentangled guidance sampling method to ensure a close alignment between the generated appearance of person image and the target pose, thereby guaranteeing the authentic restoration of details during pose transformations.
	
\end{itemize}

\section{Proposed methods}
\label{sec:tech}

\begin{figure*}[bht]
\centering
\includegraphics[width=160mm,height=58mm]{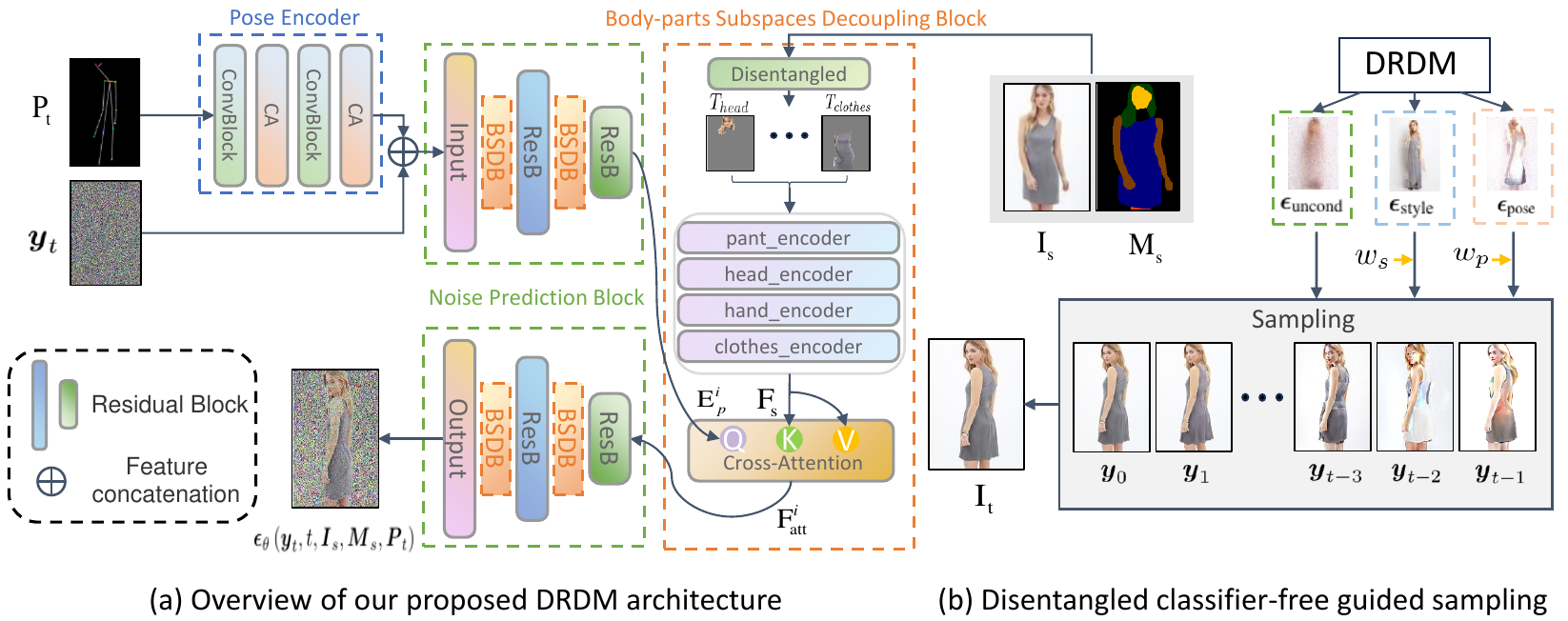}
\vspace{-2.5 mm}
\caption{Overview of our model. (a) DRDM consists of the BSDB, pose encoder $E_{p}$, and noise prediction block $\mathcal{H}_{N}$.
(b) We design a disentangled classifier-free guided sampling, which amplifies the conditional signals of texture and pose based on parsing map, ensuring a close alignment between the appearance of the generated person image and the target pose.}
\vspace{-4.95 mm}
\label{fig:02}
\end{figure*}

\textbf{Overall Framework:} Our model focuses on achieving feature-decoupled person image synthesis, enabling realistic persona generation by flexibly manipulating the appearance and body pose of the source person image.
This method excels in addressing common occlusion challenges, severe distortion at limb-end and deviation at cloth style within this task.
The overall model architecture is illustrated in Figure~\ref{fig:02}.
The DRDM framework consists of the BSDB, pose encoder $E_{p}$, and noise prediction block $\mathcal{H}_{N}$.
During the generation of the target person image, we employ $E_{p}$ to encode the input pose $P_{t}$ into a high-dimensional space, which guides the network in recombining the decoupled texture features of the source person image.
We employ a BSDB to decouple features of different body parts in the source person image $I_{s}$ to address occlusion and limb-end distortion issues.
The disentangled textures of the various body parts are separately encoded and stacked to form a multi-scale disentangled texture feature $F_{s}$.
Subsequently, the BSDB uses a cross-attention mechanism guided by the target pose to extract the required texture style from $F_{s}$ and inject it into different layers of $\mathcal{H}_{N}$.
This process enables the network to generate high-fidelity texture details.
Decoupling the textural features of the source person image helps the network in learning the relationship between body part textures and human structure, facilitating effective resolution of common issues, such as occlusion and distortion at limb-end extremities.
Additionally, we can explicitly manipulate the appearance and body pose of the source portrait by replacing body part textures and body poses through the disentangled feature representation.
During the inference, we propose a classifier-free disentangled guided sampling method to amplify the conditional signals of the disentangled textures and the target pose.
This approach enables us to closely align the generated person image’s appearance with the target pose to enhance the quality of the generated images.

\subsection{Pose Encoder}
The proposed pose encoder $E_{p}$ is composed of multiple channel attention (CA) layers and convolutional down-sampling layers.
A pre-trained human pose estimation method~\cite{cao2017realtime} is employed to obtain estimated joint positions from the human body pose and construct an 18-channel pose estimation map as the pose input $P_{t}$.
$E_{p}$ encodes $P_{t}$ into a high-dimensional space to guide the network in reassembling the disentangled texture features of the source person image, thus achieving better texture and pose transfer effects.

\subsection{BSDB}
To address occlusion and limb-end distortion issues, we propose the BSDB.
BSDB utilizes the semantic segmentation map $M_{s}$ to separate the features of the different body parts in the source person image $I_{s}$, achieving the disentanglement of features for the various semantic regions of the person.
These disentangled features of different body parts are represented as ${T}=[{T}_{head},{T}_{cloth},{T}_{pant},{T}_{hand}]$.
To further achieve feature disentanglement, we pass each disentangled body part feature through the corresponding texture encoder and stack the outputs to obtain the multi-scale feature representation ${F}_{s}$ for the entire source person image.
A self-attention within the texture encoders for each body part is introduced to address the issue of texture distortion in generated results due to inaccuracies in semantic segmentation maps.
Which ensures that the encoders solely focus on the texture of the respective body part they are responsible for, rather than encoding the textures of other body parts together.
This mechanism exploits the parsing map as a regularization factor for preserving the figure shape and body part subspace texture style, which guarantees the isolation of texture encoding for each individual body part, thereby alleviating the texture distortion problem caused by inaccuracies in semantic segmentation.
Finally, we use cross-attention mechanisms at resolutions of $32\times32$, $16\times16$, and $8\times8$ to provide the required features to the noise prediction block ($\mathcal{H}_{N}$).
At each resolution, we use network outputs containing pose features ${E}^{i}_{p}$ to extract the required texture style from $F_{s}$ and inject it into different layers of $\mathcal{H}_{N}$.
Decoupling the features of the source person image helps the network in learning the relationship between body part textures and human structure, enabling effective resolution of common issues, such as occlusion and distortion at limb-end extremities.

\subsection{Noise Prediction Block}
We use the pose ${P}_{t}$ to guide the noise prediction block ($\mathcal{H}_{N}$) in predicting the noise ${\epsilon}_{\theta}\left({y}_{t}, t, {I}_{s}, {M}_{s}, {P}_{t}\right)$ throughout the entire process of noise prediction.
During this phase, the BSDB employs cross-attention mechanisms to provide the required multi-scale texture features $F_{s}$ to $\mathcal{H}_{N}$, thereby aiding $\mathcal{H}_{N}$ in more accurately predicting the noise.
In the training process, we generate noise samples $\boldsymbol{y}_{t} \sim q\left(\boldsymbol{y}_{t} \mid \boldsymbol{y}_{0}\right)$ by adding random Gaussian noise $\epsilon$ to $y_{0}$ to train our network.



\subsection{Parsing Map-based Disentangled Guided Sampling}
To achieve higher image quality, we design a disentanglement guided sampling approach based on classifier-free guidance~\cite{ho2021classifier} and parsing map, named parsing map-based disentanglement classifier-free guidance (PMDCF-guidance) sampling.
In practice, we can iteratively sample $\boldsymbol{y}_{T}$ from $p_{\theta}\left(\boldsymbol{y}_{t-1} \mid \boldsymbol{y}_{t},{I}_{s},{M}_{s},{P}_{t}\right)$ from $t=T$ to $t=1$, converging to the learned conditional and unconditional distributions.
The parameterization for predicting the mean and variance of $\boldsymbol{y}_{t-1}$ is as follows:

\vspace{-4.5 mm}
\begin{eqnarray}
p_{\theta}(\boldsymbol{y}_{t-1} \mid \boldsymbol{y}_{t}, {I}_{s}, {M}_{s}, {P}_{t}) =\qquad\qquad\nonumber \\
\mathcal{N}(\boldsymbol{y}_{t-1};\mu_{\theta}(\boldsymbol{y}_{t},t,{I}_{s},{M}_{s},{P}_{t}),	
\Sigma_{\theta}\left(\boldsymbol{y}_{t}, t, {I}_{s}, {M}_{s}, {P}_{t}\right)).
\end{eqnarray}
\vspace{-4.5 mm}

In the process, we effectively decouple and amplify the conditional signals of texture and pose, improving the alignment of the generated person image’s appearance with the target pose and enhancing the quality of the generated images.
The formulation is as follows:
\vspace{-4.5 mm}

\begin{eqnarray}
	\boldsymbol{\epsilon}_{\text {cond }} & = & \boldsymbol{\epsilon}_{\text {uncond }} + w_{s}\boldsymbol{\epsilon}_{\text {style }} + w_{p}\boldsymbol{\epsilon}_{\text {pose}},
\end{eqnarray}
where, $\epsilon_{\mathrm{uncond}}=\epsilon_{\theta}\left(\boldsymbol{y}_{t}, t, \emptyset, \emptyset,\emptyset\right)$ represents the model's unconditional prediction.
The style-guided prediction is denoted as $\epsilon_{\mathrm{style}} = \epsilon_{\theta}\left(\boldsymbol{y}_{t}, t, \boldsymbol{I}_{s},\boldsymbol{M}_{s},\boldsymbol{P}_{t}\right)-\epsilon_{\theta}\left(\boldsymbol{y}{t}, t, \emptyset,\emptyset,\boldsymbol{P}_{t}\right)$, and the pose-guided prediction is indicated as $\epsilon_{\mathrm{pose}} = \epsilon_{\theta}\left(\boldsymbol{y}_{t}, t, \emptyset,\emptyset,\boldsymbol{P}_{t}\right)-\epsilon_{\theta}\left(\boldsymbol{y}_{t}, t, \emptyset, \emptyset,\emptyset\right)$.
Where, $\emptyset$ is a zero tensor used to replace the missing conditions, $w_{s}$ and $w_{p}$ are the guidance scale for style conditions and pose conditions repectively.

\section{Experiments}
\label{sec:experiments}

\textbf{Experimental Details:} We evaluate the performance of DRDM by conducting experiments on the widely used DeepFashion dataset,utilizing the Rebalanced Parsing model~\cite{enbo2020rebalance} to generate parsing maps.
During the training phase, the Adam optimizer is configured with a learning rate of $2e-5$, and the batch size is $7$.
We randomly set 10$\%$ of variables ${I}_{s}$, ${M}_{s}$, and ${P}_{t}$ to zero tensors to encourage the model to effectively learn the unconditional distribution.
For sampling, we set $w_{s}$ to 3.0 and $w_{p}$ to 2.5.

Our experiments were conducted on a system equipped with four NVIDIA V100 GPUs.

\subsection{Quantitative Comparison}
We conduct a quantitative comparison of our proposed DRDM with several state-of-the-art methods~\cite{bhunia2023person,ren2022neural,zhou2022cross,zhang2021pise,ren2020deep,men2020controllable}.
Table~\ref{Quantitative evaluation} displays the results, showing that we achieved the best scores in terms of SSIM and LPIPS metrics.

\begin{table}[h]
\caption{Quantitative comparison of the proposed DRDM with several state-of-the-art models in metrics of SSIM, FID and LPIPS.}
\vspace{-1.5mm}
\label{Quantitative evaluation}
\centering
\begin{tabular}{lp{21mm}<{\centering}m{15mm}<{\centering}m{15mm}<{\centering}m{15mm}<{\centering}}
\hline
\rule{0pt}{10pt}Methods  & FID (↓)   & SSIM (↑) & LPIPS (↓)\\
\hline\hline
\rule{0pt}{9pt}Def-GAN    & 18.457  & 0.6786  & 0.2330     \\
\rule{0pt}{9pt}PATN       & 20.751  & 0.6709  & 0.2562     \\
\rule{0pt}{9pt}ADGAN      & 14.458  & 0.6721  & 0.2283     \\
\rule{0pt}{9pt}PISE       & 13.610  & 0.6629  & 0.2059     \\
\rule{0pt}{9pt}GFLA       & 10.573  & 0.7074  & 0.2341     \\
\rule{0pt}{9pt}DPTN       & 11.387  & 0.7112  & 0.1931     \\
\rule{0pt}{9pt}CASD       & 11.373  & 0.7248  & 0.1936     \\
\rule{0pt}{9pt}NTED       & 8.6838  & 0.7182  & 0.1752     \\
\rule{0pt}{9pt}PIDM       & 7.8615  & 0.7312  & 0.1678     \\
\hline
\rule{0pt}{11pt}\pmb{DRDM (Ours)} & \pmb{7.7462}  & \pmb{0.7409}  & \pmb{0.1652}      \\
\hline
\end{tabular}
\vspace{-4.5 mm}
\end{table}

This notion means that our method can preserve complete and clear textural structures even under perfect alignment of the target pose.
In fact, GAN methodologies may produce conspicuous artifacts when confronting significant texture deformations.
Meanwhile, the diffusion-based PIDM~\cite{bhunia2023person} frequently struggles to retain clothing styles and limb details in cases with occlusions.

\begin{figure}[ht]
\centering
\includegraphics[width=88mm,height=77mm]{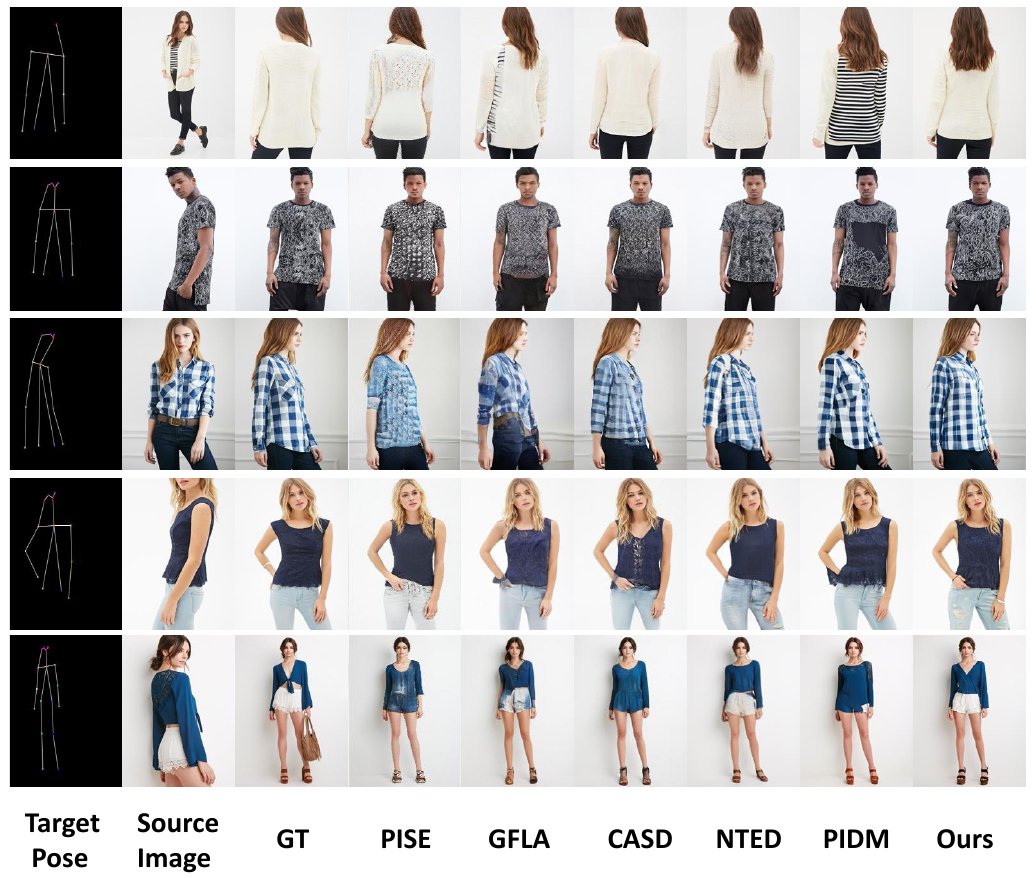} 
\vspace{-6.5 mm}
\caption{Qualitative comparisons with several state-of-the-art models on the DeepFashion dataset.} 
\label{fig:03}
\end{figure}

Five representative results are shown in Figure~\ref{fig:03}.
The composition images for the baseline methods were generated using pre-trained models provided by their respective authors.
Notably, our model demonstrates superior performance in preserving fine texture details, maintaining accurate forms, and reconstructing realistic limbs and facial features.

\vspace{-3.5 mm}
\vspace{-3.5 mm}

\subsection{User Study}

While quantitative comparisons offer valuable insights into various aspects of the generated results, human image composition tasks are often user-centric.
Accordingly we conduct a user study involving 90 volunteers to evaluate the perceptual performance of the generated results.
The user study consisted of two parts.
(a) Comparison with real images: We randomly select 90 real images and 90 generated images from the test set and shuffled them.
The volunteers are required to determine whether the given image is real or synthetic within 1 second.
(b) Comparison with other methods: We present 90 pairs of randomly selected images to the volunteers, including source images, target poses, real images, images generated by our method, and baseline methods.
The volunteers are asked to choose the most realistic and plausible image compared with the source and real images.
In the user study, R2G represents the percentage of real images mistaken for generated images, G2R denotes the percentage of generated images mistaken for real images, and Jab indicates the percentage of images judged as the best among all models.
Our method exhibits the best results, indicating that the results generated by our method are more aligned with human perception, as shown in Table~\ref{usr evaluation}.

\begin{table}[h]
\vspace{-2.5 mm}
\caption{Exhibition of user study in metrics of R2G, G2R and Jab, computed from users’ feedback shown on $256\times176$ resolution for DeepFashion.}
\vspace{-2.5 mm}
\label{usr evaluation}
\centering
\begin{tabular}{lp{21mm}<{\centering}m{14mm}<{\centering}m{14mm}<{\centering}m{15mm}<{\centering}}
\hline
\rule{0pt}{10pt}Methods & R2G (↑) & G2R (↑) & Jab (↑)\\
\hline\hline
\rule{0pt}{9pt}ADGAN &  22.2$\%$      &   22.2$\%$     &   4.22$\%$    \\
\rule{0pt}{9pt}PISE &  29.2$\%$      &   25.8$\%$     &   5$\%$    \\
\rule{0pt}{9pt}GFLA &  23$\%$     &    26.2$\%$    &  5.77$\%$     \\
\rule{0pt}{9pt}CASD &  34.7$\%$      &    31.7$\%$    &   5.77$\%$    \\
\rule{0pt}{9pt}NTED &  32.4$\%$      &    36$\%$ &   13$\%$    \\
\rule{0pt}{9pt}PIDM &    36.3$\%$    &    43$\%$    &    20.4$\%$   \\
\hline
\rule{0pt}{9pt}\pmb{DRDM (Ours)} &   \pmb{40.7$\%$}      &    \pmb{51.4$\%$}     &    \pmb{45.7$\%$}   \\
\hline
\end{tabular}
\vspace{-3.5 mm}
\end{table}

\subsection{Ablation Study}
We further evaluate our proposed method by means of ablation experiments.
We created three variants by selectively removing specific components from the complete model (w/o BSDB and w/o self-attention) and by not using our proposed sampling method (w/o PMDCF-guidance).
Table~\ref{ablation evaluation} illustrates that our complete model achieved the best performance in all quantitative comparisons.
The results highlight the significance of each component in contributing to the overall performance of the final outcomes.

\vspace{-2.5 mm}
\begin{table}[h]
\caption{Ablation studie on DeepFashion dataset.}
\vspace{-0.4mm}
\label{ablation evaluation}
\centering
\begin{tabular}{lp{21mm}<{\centering}m{13mm}<{\centering}m{14mm}<{\centering}m{15mm}<{\centering}}
\hline
\rule{0pt}{10pt}Methods             & FID (↓)   & SSIM (↑) & LPIPS (↓) \\
\hline\hline
\rule{0pt}{9pt}w/o BSDB            & 10.598   & 0.7016  & 0.2487  \\
\rule{0pt}{9pt}w/o self-attention  & 7.9172   & 0.7365  & 0.1733  \\
\rule{0pt}{9pt}w/o PMDCF-guidance    & 9.1425   & 0.7366  & 0.1747  \\
\hline
\rule{0pt}{11pt}\pmb{Model-full (ours)}    & \pmb{7.7462}  & \pmb{0.7409}  & \pmb{0.1652}  \\
\hline
\end{tabular}
\end{table}

\section{Conclusion}
\label{sec:conclusion}
In this work, we proposed a DRDM that allows for a precise control of the pose and appearance of the source person image to synthesize realistic person images.
In practice, there are three types of degeneration key issues existing in portrait composition, \emph{i.e.}, details missing, limbs distortion and garment style deviation.
We exploit the parsing map as a regularization factor for preserving the figure shape and body part subspace texture style, and then extract structure and texture style features in each subspace.
The introduction of the disentangled texture fusion block enables the decoupling of textures in different semantic parts of the source person image, which can then be guided by the target pose to compose the appearance of the target person.
Additionally, we utilize a parsing map-based disentangled classifier-free guided sampling approach to achieve accurate detail reconstruction under pose variations.
We demonstrate the effectiveness of DRDM for controllable person image synthesis task through extensive quantitative, ablation and user studies.

\bibliographystyle{IEEEbib}
\bibliography{main}

\end{document}